\documentclass{article}
\usepackage{ijcai18}

\usepackage{times}
\usepackage{xcolor}
\usepackage{soul}
\usepackage[small]{caption}

\usepackage{latexsym} 

\usepackage{graphicx}
\usepackage{bm}
\title{Predicting the Popularity of Online Videos via Deep Neural Networks}

\author{
Yue Mao, 
Yi Shen, 
Gang Qin,
Longjun Cai
\\
Alibaba Group\\
\{maoyue.my, sy133447, gary.qg, longjun.clj\}@alibaba-inc.com
}

\begin{document}

\maketitle

\begin{abstract}
	Predicting the popularity of online videos is important for video streaming content providers. This is a challenging problem because of the following two reasons. First, the problem is both ``wide'' and ``deep''. That is, it not only depends on a wide range of features, but also be highly non-linear and complex. Second, multiple competitors may be involved. In this paper, we propose a general prediction model using the multi-task learning (MTL) module and the relation network (RN) module, where MTL can reduce over-fitting and RN can model the relations of multiple competitors. Experimental results show that our proposed approach significantly increases the accuracy on predicting the total view counts of TV series with RN and MTL modules. 
\end{abstract}

\section{Introduction}
Video streaming has become one of the most popular online activities in recent years. Today's video streaming market has become highly competitive and it has been crowded with various content providers. For example, in China, there are Youku, IQiyi, Sohu, LeTV and Tencent Video, etc. In North America, there are Netflix, Hulu and Amazon Video, etc. For these content providers, predicting the future video popularity is very important, for example, in planning advertisements. Although there are many types of contents such as movie, TV series, drama, cartoon, MTV and other user generated contents, we focus on predicting the popularity of TV series in this paper. One of the most used measurements of online video popularity is the view count. Suppose a TV series will be released on Youku\footnote{http://www.youku.com} in $7$ days, is it possible to predict the total view count of all episodes of this TV series? Generally, this problem is challenging for the following reasons.

First, the prediction problem is both ``wide'' and ``deep''. Similar to the wide and deep learning for recommender systems \cite{widedeep}, the prediction problem not only depends on a wide range of features but also be highly non-linear and complex. Modern deep learning provides a very powerful framework for supervised learning. With more layers and units, deep neural network can represent complex and highly nonlinear models. Also, deep neural networks perform better in generalizing to unseen feature combinations.  However, over-fitting becomes a serious issue when the data available is limited. Multi-task learning (MTL) \cite{MTL-original}\cite{MTL97} is one way to improve the generalization power of the original task, by jointly training multiple related tasks and sharing representations between them. Multi-task learning is widely used with success in machine learning, and it acts as a regularizer and it reduces the risk of over-fitting \cite{MTL-overfitting} and its ability to fit random noise. In addition to MTL, dropout \cite{dropout} and $\ell_2$ regularization are also useful to prevent over-fitting.

Second, multiple competitors are involved in the prediction problem. Users often migrate among different content providers for many possible reasons \cite{migrate}. To predict a TV series's video views, one must consider the influence of other related TV series from various content providers. Some special structure should be involved in the prediction model to represent the relations of multiple competitors. Then relation networks (RN) introduced in \cite{RN} become useful, where originally a visual reasoning problem is considered. An RN is a simple and effective module for the case when input ``objects'' (usually hidden units of a neural network) have relations which each other. Relation network is also found to be useful in neural machine translation \cite{RN-nmt}, few-shot learning \cite{rn-fewshot} and complex relational reasoning \cite{RN-recurrent}. Relation networks are also related to relational learning in statistics \cite{statRelational}. 

In this paper, we introduce a general model for predicting the popularity of online videos. This paper is organized as follows.  In Section \ref{sec_back}, we give background knowledge on multi-task learning and relation networks. In Section \ref{sec_prop}, we give the details of our proposed prediction model with the DNN module, the RN module and the MTL module. The experimental results are discussed in Section \ref{sec_experiment} and the related work is discussed in Section \ref{sec_related}. 

\section{Background} \label{sec_back}
In this section, we briefly introduce multi-task learning and relation networks, which are used as separate modules in our prediction model.

\subsection{Multi-task learning}
\begin{figure}[h]
	\begin{center}
		\includegraphics[width=0.35\textwidth]{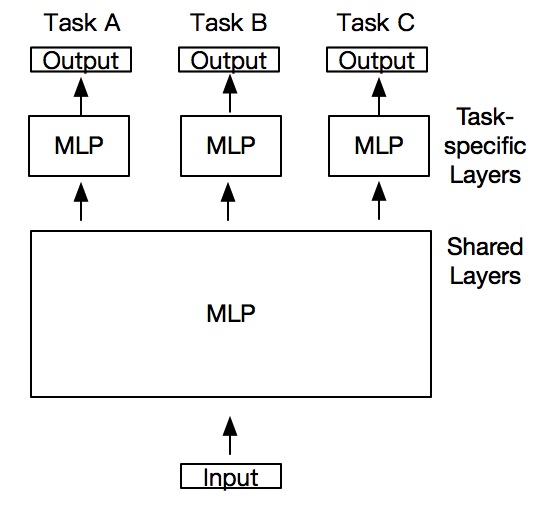}
	\end{center}
	\caption{Hard parameter sharing for multi-task learning in deep neural networks.}\label{fig-MTL}
\end{figure}

Multi-task learning is typically done by hard or soft parameter sharing. In this paper, we use hard parameter sharing. It shares the hidden layers between all tasks, while keeps several task-specific output layers \cite{MTL}. The idea is shown in Figure \ref{fig-MTL}.

Suppose $\mathcal{L}$ is the loss for the main task and $\mathcal{L}'_i$s are the losses for some other auxiliary tasks, then they can be trained jointly with the total loss
\begin{equation}
\mathcal{L}_{total} = \mathcal{L} + \sum_{i}\lambda_i \mathcal{L}_i
\end{equation}
where $\lambda_i$ is the weight for the $i$-th auxiliary task. The value of $\lambda_i$ is used to control the importance of an auxiliary task.

\subsection{Relation network}
Given a set of ``objects'' $\{\bm{o_1},\bm{o_2},...,\bm{o_n}\}$, which are hidden units of a neural network. A relation network is defined as
\begin{equation} \label{eqnRN1}
f\left(\sum_{1 \leq i < j \leq n}g (\bm{o_i},\bm{o_j})\right)
\end{equation}
where $\bm{o_i}$ is the $i$-th object and $f$ and $g$ are differentiable functions. Here, the pair-wise relations are represented by common function $g$ and the combined relation of $\{\bm{o_1},\bm{o_2},...,\bm{o_n}\}$ is fed to the function $f$. 

\begin{figure}[h]
	\begin{center}
		\includegraphics[width=0.4\textwidth]{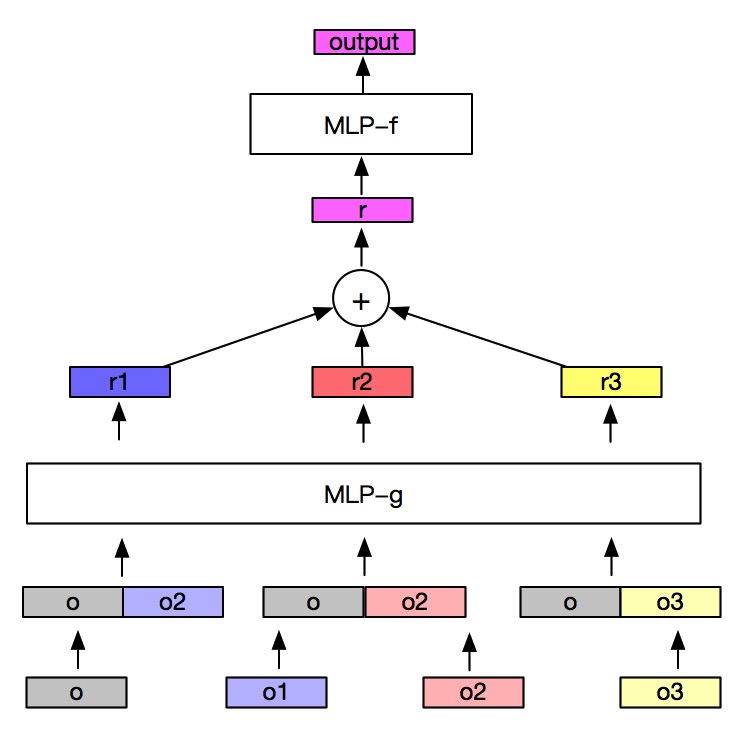}
	\end{center}
	\caption{The relation network to represent relation of $\bm{o}$ with $\bm{o_1}$, $\bm{o_2}$, ..., $\bm{o_n}.$}\label{fig-RN}
\end{figure}

Our prediction problem is slightly different. Suppose we want to predict one particular object, say $\bm{o}$. Instead, we consider the relations of $\bm{o}$ with the other objects $\bm{o_1},\bm{o_2},...\bm{o_n}$. Our relation network is defined by modifying (\ref{eqnRN1}) as
\begin{equation} \label{eqnRN2}
f\left( \sum_{i=1}^n g (\bm{o},\bm{o_i}) \right)
\end{equation}
and it is shown in Figure \ref{fig-RN}.

\section{Proposed Model}\label{sec_prop}
\begin{figure}[h]
	\begin{center}
		\includegraphics[width=0.45\textwidth]{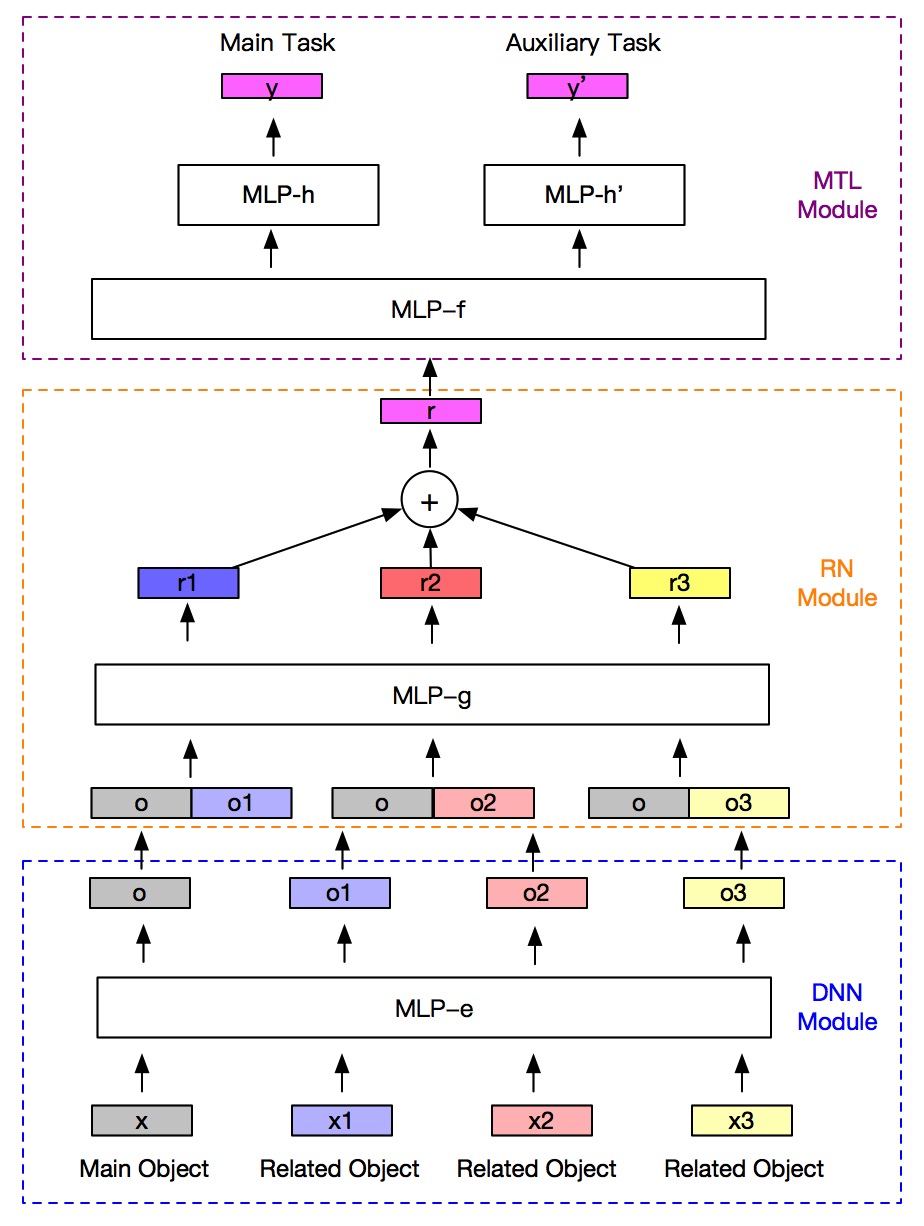}
	\end{center}
	\caption{Our proposed prediction model. It consists of three modules:  the deep neural network (DNN) module, the relation network (RN) module and the multi-task learning (MTL) module.}\label{fig-model}
\end{figure}

The architecture of our proposed model is shown in figure \ref{fig-model}. Our model contains the deep neural network (DNN) module, the relation network (RN) module and the multi-task learning (MTL) module. We will describe the details in this section.  

\subsection{Problem definition}
We define our problem mathematically. The input vector representing features is denoted as $\bm{x}$. For notation simplicity, we assume only one auxiliary task in our MTL module is jointly trained, and the corresponding outputs of the main task and the auxiliary task are denoted as $y$ and $y'$, respectively. In practice, more than one auxiliary tasks may be trained jointly depends on the problem given. 

Let $(\bm{x},y,y')$ denote the input vector, the main task output and the axillary task output. Suppose our training dataset is 
\begin{equation}
\mathcal{D} = \{(\bm{x},y,y')\}
\end{equation}
 Assume any sample $\bm{x} \in \mathcal{D}$ has relation with some other $n$ samples in $\mathcal{D}$, say $\bm{x_1},\bm{x_2},..,\bm{x_n} \in \mathcal{D}$ where $n$ is a fixed integer. Then the original training dataset $\mathcal{D}$ can be transformed to a new dataset 
\begin{equation}
\mathcal{{D}}^* = \{(\bm{x^*}, y, y')\}
\end{equation}
where $\bm{x^*} = concat(\bm{x},\bm{x_1},\bm{x_2},..,\bm{x_n})$. We call these related samples $\bm{x},\bm{x_1},\bm{x_2},..,\bm{x_n}$ as ``objects'', which coincide with the terminology used in relation networks \cite{RN}.
\subsection{Our prediction model}
Consider the main input vector $\bm{x}$ (main object) and the related input vectors $\bm{x_i}$ (related objects), for $i=1,2,...,n$, where $n$ is a fixed integer corresponding to $n$ related objects. The value of $n$ depends on the specific problem given. For instance, if we want to predict the popularity of a TV series, then $n$ related TV series should be considered, usually popular TV series with the closest releasing dates. 

The first part is the DNN module. We use deep multi-layer perceptrons (MLPs) to get their high-level representations
\begin{equation}
\bm{o}= e(\bm{x}),\quad \bm{o_i}=e(\bm{x_i}),\quad i=1,2,...,n.
\end{equation}
We use many layers of MLPs for the function $e$ to capture the complex combinations of features (more than $10$ layers). 
 
The next part is the RN module. Let the function $f$ and $g$ be two shallow MLPs (about $2$ or $3$ layers). The relation of $\bm{o}$ and $\bm{o_i}$ is represented by the function $g$,
\begin{equation}
\bm{r_i} = g(\bm{o},\bm{o_i}),\quad i=1,2,...,n.
\end{equation}
The combined relation is fed to $f$ as an input,
\begin{equation}
\bm{r} = f(\sum_{i=1}^n \bm{r_i})
\end{equation}
where the summation is element-wise.

The last part is the MTL module. The vector $\bm{r}$ from RN module is fed to the MTL module as a input. Note the shared dense layers of MTL in Figure \ref{fig-MTL} are skipped here, as there are already MLPs in the RN module.  Therefore, the vector $\bm{r}$ is directly fed to two separate task-specific MLPs, say $h$ for the main task and $h'$ for the auxiliary task. Finally,  we get the main output and the auxiliary output, respectively, 
\begin{equation}
\hat{y} = h(\bm{r}),\quad \hat{y}' = h'(\bm{r}).
\end{equation}

In conclusion, for any input vector $(\bm{x^*},y,y')$ from the training dataset $\mathcal{D}^*$, we output $\hat{y}(\bm{x^*},\bm{\theta})$ and $\hat{y}'(\bm{x^*},\bm{\theta})$. The loss function is
\begin{eqnarray}
\mathcal{L}(\mathcal{D}^*,\bm{\theta}) &= &\sum_{(\bm{x^*},y,y') \in \mathcal{D}^*} |\hat{y}(\bm{x^*},\bm{\theta})-y|^2 \nonumber \\ 
& &+\lambda\sum_{(\bm{x^*},y,y') \in \mathcal{D}^*}|\hat{y}'(\bm{x^*},\bm{\theta})-y'|^2 \nonumber \\ 
& & +\gamma || \bm{\theta}||_2^2 
\end{eqnarray}
where $\lambda$ and $\gamma$ are hyper-parameters for MTL and $\ell_2$ regularization, respectively.

\section{Experiments} \label{sec_experiment}
In this section, we discuss the experimental results on predicting the total view counts of TV series on Youku. We use more than 200 dimensional features as our input, including the actor/actress name, genre and director name, etc. 

For the DNN module, we use $15$ layers of MLPs. For the RN module, we fix the number of related objects as $n=3$. We use $3$ layers of MLPs to represent a ``relation'' and the combined relation is fed to another $3$ layers of MLPs.  For the MTL module, we jointly predict the total view count and the popularity index of Youku, which represents the TV series' popularity. 

The ReLU activation is used to provide the non-linearity. We also use batch normalization \cite{BN} to prevent gradients vanishing, and dropout \cite{dropout} and $\ell_2$ regularization to reduce over-fitting and increase the model's generalization power. Our model is end-to-end differentiable. The loss function is optimized by back propagation and accelerated with Momentum, using our self-developed JAVA framework of deep learning called DeepDriver \footnote{http://github.com/LongJunCai/DeepDriver}.

Note that the training dataset vary in time, hence, the date when the prediction is made is important. Typically, we predict $7$, $180$ and $360$ days before the TV series' releasing date. To evaluate the effectiveness of our prediction model, we compare the full model ``DNN+RN+MTL'' with the partial models ``DNN'' and ``DNN+MTL''. All TV series before Jan 1, 2017 are used as training data, and the other TV series from Jan 1, 2017 are used for evaluating. 

The results are shown in terms of coefficients of determination ($\mathcal{R}^2$) in Table \ref{tableResult}. For example, the $\mathcal{R}^2$ predicted $7$ days before the releasing date, the result is improved by $0.03$ with the MTL module and another $0.04$ with the additional RN module.

\begin{table}[h] 
	\begin{center}
		\begin{tabular}{|c|c|c|c|}
			\hline
			&DNN&DNN+MTL&DNN+RN+MTL\\
			\hline
			7 days&  $0.80$& $0.83$ & $0.87$ \\
			\hline
			180 days&  $0.75$& $0.80$ & $0.85$\\
			\hline
			360 days& $0.74$& $0.79$ & $0.85$\\
			\hline
		\end{tabular}
		\caption{The results of predicting future view counts of TV series are shown in terms of coefficients of determination ($\mathcal{R}^2$). It is trained on TV series data before Jan 1, 2017 and evaluated on TV series data from Jan 1, 2017. The first, second and third row corresponds to predictions results made $7$, $180$ and $360$ days before the TV series' releasing date, respectively. }\label{tableResult}
	\end{center}
\end{table} 

\section{Related Work} \label{sec_related}
Much work has been done for predicting the popularity of web content, a survey is done in \cite{surveyPred}. In particular, there are a lot of approaches for predicting the popularity of online videos. In \cite{predSVR}, the popularity of Youtube and Facebook videos is predicted using support vector regression with Gaussian radial basis functions. In \cite{predSentiment}, video popularity is predicted with sentiment propagation via implicit network.  A dual sentimental Hawkes process (DSHP) is proposed, and it is evaluated on four types of videos: movies, TV episodes, music videos, and online news. In \cite{predYoutube}, four methods are discussed for predicting the popularity of Youtube videos: univariate linear model, multivariate linear model, radial basis functions and a preliminary classification method. In \cite{predYouku}, both online video future popularity level and online video future view count are predicted for Youku videos. In \cite{predYouku2}, how the popularity of Youku videos evolved over time is analyzed, and multivariate regression models with different parameters are used. In \cite{predUGC}, predicting the popularity of user-generated contents (UGC) is discussed. In \cite{predSerial}, the popularity of online serials is predicted with autoregressive models, including naive auto-regressive (NAR) model based on the correlations of serial episodes, and transfer auto-regressive (TAR) model to capture the dynamic behaviors of audiences. In \cite{predFoursquare}, multivariate linear regression is used to predict the micro-review popularity using the location-based social networking platform Foursquare.

The ``wide'' and ``deep'' prediction problem is similar to analyzing the high dimension low sample size (HDLSS) data. Feature selection is a powerful and widely used tool for HDLSS data. For instance, Lasso \cite{lasso} minimizes the objective function penalized by the $\ell_1$ norm and leads a sparse model. Unfortunately, Lasso considers only the linear dependency but ignores the nonlinearity. In \cite{dnp}, deep neural pursuit (DNP) is proposed with advantages in high nonlinearity, the robustness to high dimensionality, the capability of learning from a small number of samples. In \cite{widedeep}, a Wide\&Deep model is proposed for recommender system and it achieves both memorization and generalization. Wide linear models can effectively memorize sparse feature interactions using cross-product feature transformations, while deep neural networks can generalize to previously unseen feature interactions through low-dimensional embeddings. Our work is related to the above methods, but we extend the problem by considering the competition relation of objects, which is very common in practice.

\section{Conclusion}
In this paper, we propose a model for predicting the popularity of online videos via deep neural networks with MTL and RN modules. Our model can apply to situations where the problem is both ``wide'' and ``deep'', and with multiple competitions. The experimental results on predicting the total view counts of TV series suggest our model is effective. 

In the future work, we plan to improve our model with recurrent neural networks (RNN) related structures such as long short-term memory (LSTM) networks \cite{lstm}. Approaches in time series prediction may be also helpful to improve our model. Furthermore, we think even more features can be added to our current inputs, e.g., video frame information from the trailer and text information from video description, etc.  

Besides online videos, our work can be used to predict other types of web content. Our work can even apply to other similar applications, such as predicting stock prices and predicting the winner of a presidential election, etc. 

\newpage
\bibliography{cite.bib}

\end{document}